\documentclass[sigconf]{acmart}
\AtBeginDocument{%
  }

\setcopyright{none}
\copyrightyear{2026}
\acmYear{2026}
\acmConference[ICCAD '26]{2026 ACM/IEEE International Conference on Computer-Aided Design}{November 8--12,
  2026}{San Jose, California, USA}

\usepackage{multirow}
\usepackage{tabularx}
\usepackage{algorithm}
\usepackage{algorithmic}

\begin{document}

\title{LLM-Guided Safety Agent for Edge Robotics with an ISO-Compliant Perception-Compute-Control Architecture}

\author{Xu Huang$^*$, Ruofan Zhang$^*$, Lu Cheng, Yuefeng Song, Xu Huang, Huayu Zhang, Sheng Yin, Anyang Liang, Chen Qian, Yin Zhou, Xiaoyun Yuan, Yuan Cheng$^\dagger$}

\affiliation{%
  \mbox{
  \institution{Shanghai Jiao Tong University} \quad  \institution{SIMMIR Tech}
  }
  \country{}
}

\email{hxnbnnn@gmail.com}

\thanks{$*$ These authors contributed equally to this work.}
\thanks{$\dagger$ Corresponding author.}

\renewcommand{\shortauthors}{Huang et al.}


\begin{abstract}
Ensuring functional safety in human-robot interaction is challenging because AI perception is inherently probabilistic, whereas industrial standards require deterministic behavior. We present an LLM-guided safety agent for edge robotics, built on an ISO-compliant low-latency perception-compute-control architecture.
Our method translates natural-language safety regulations into executable predicates and deploys them through a redundant heterogeneous edge runtime. For fault-tolerant closed-loop execution under edge constraints, we adopt a symmetric dual-modular redundancy design with parallel independent execution for low-latency perception, computation, and control.
We prototype the system on a dual-RK3588 platform and evaluate it in representative human-robot interaction scenarios. The results demonstrate a practical edge implementation path toward ISO 13849 Category 3 and PL d using cost-effective hardware, supporting practical deployment of safety-critical embodied AI.
\end{abstract}



\maketitle

\begin{figure}[htbp]
\centerline{\includegraphics[width=0.5\textwidth]{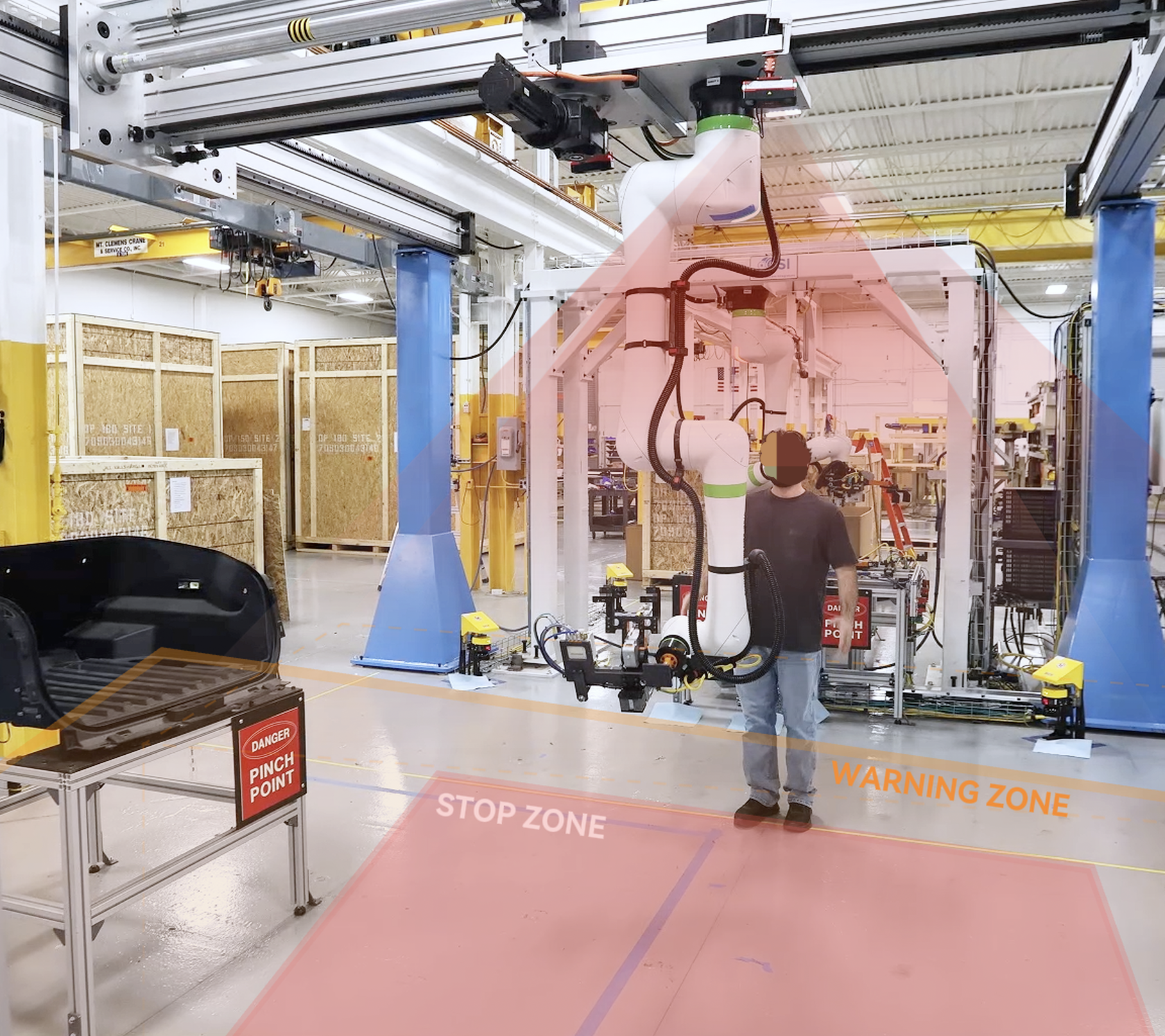}}
\caption{Real-world deployment of the proposed LLM-guided deterministic safety agent on a redundant dual-RK3588 SoC platform.}
\label{forward}
\end{figure}

\begin{figure*}[htbp]
\centerline{\includegraphics[width=1.0\textwidth]{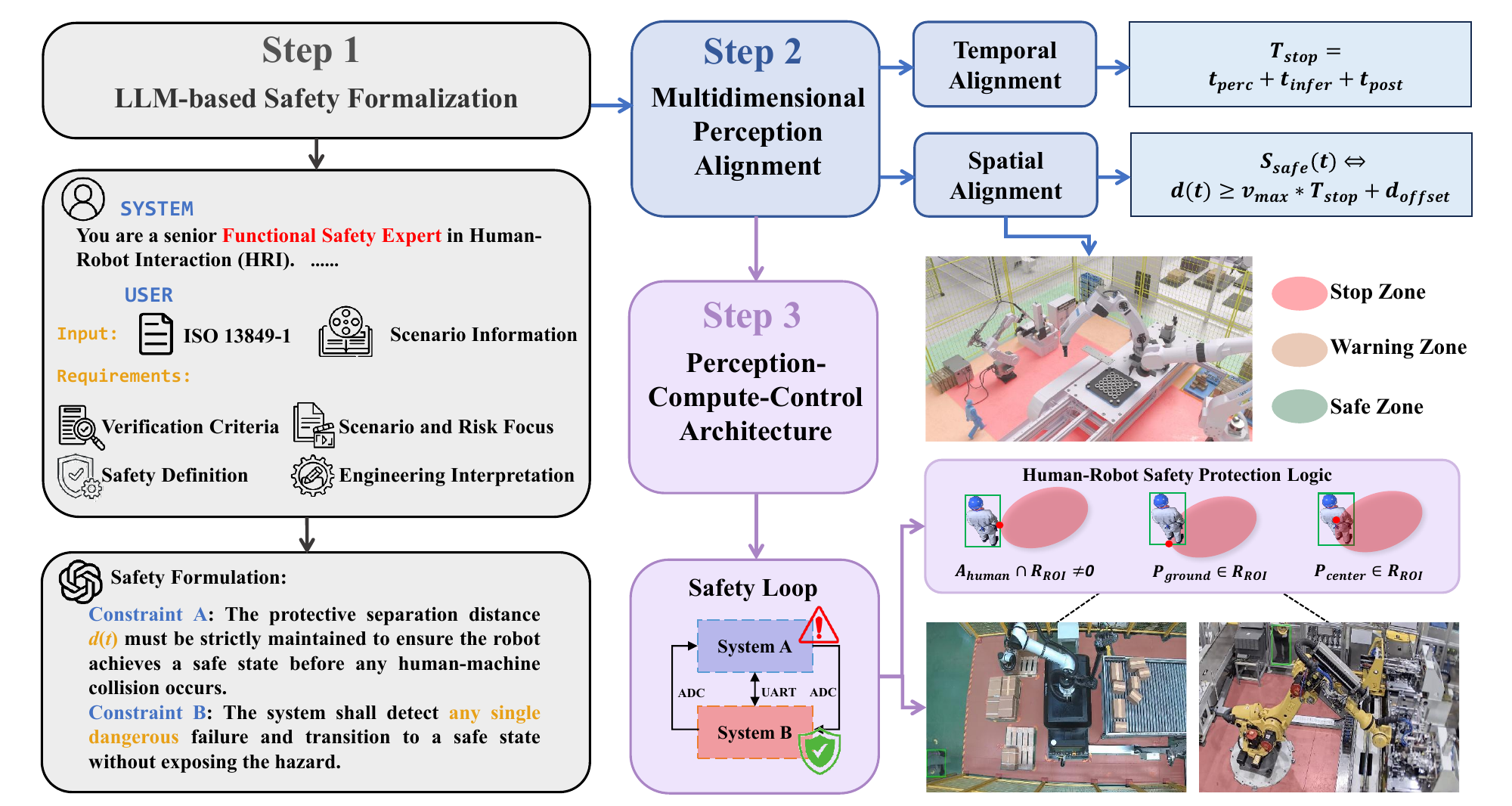}}
    \caption{The hierarchical framework of the safety agent.
    \textbf{Step 1}: LLM-based safety formalization for constraint extraction ($C_A, C_B$). 
    \textbf{Step 2}: Multidimensional perception alignment including spatial mapping ($S_{safe}$) and temporal WCET analysis ($T_{stop}$). 
    \textbf{Step 3}: The Perception-Compute-Control architecture.}
\label{framework}
\end{figure*}

\section{Introduction}

The ongoing industrial metamorphosis towards Industry 5.0 \cite{industry5.0} has prioritized human-centric flexibility, demanding a shift from static robotic isolation to dynamic Human-Robot Collaboration (HRC) \cite{HRC_review}. In these environments, human operators and robotic agents share a common physical footprint, necessitating a fundamental rethink of functional safety. However, current safety protocols often suffer from spatial reductionism: they define safety primarily through distance-based separation (e.g. ISO/TS 15066), treating the workspace as a simple binary zone \cite{rosenstrauch2018human}. Such definitions fail to account for the complexity of collaborative tasks, leading to a critical semantic bottleneck. Traditional sensors remain semantically blind \cite{brunke2025semantically}, unable to distinguish between a certified operator and an unauthorized intruder, or between a standard workflow and a hazardous anomalous action, resulting in spurious trips or latent safety risks.

To address these deficiencies, this work argues that a comprehensive HRC safety definition must transcend simple spatial proximity. We propose a multi-tiered safety semantic hierarchy that redefines functional safety across three indispensable dimensions: (i) \textbf{Human-Robot Safety Protection} (Spatial Safety), which maintains the fundamental physical boundaries; (ii) \textbf{Anomalous Target Identification} (Object-Level Integrity), which verifies the legitimacy of entities within the workspace; and (iii) \textbf{Anomalous Behavior Recognition} (Temporal Semantic Safety), which evaluates the intent and dynamics of human movement to anticipate risks before they manifest physically. To operationalize this 3D definition, we leverage Large Language Models (LLMs) as a design-time semantic bridge for semi-automated safety formalization. Under a domain-specific prompting strategy, the LLM assists in parsing complex regulatory texts (e.g., ISO 13849-1 \cite{iso13849}) and maps their abstract requirements onto this tiered hierarchy, creating a formalization pipeline that maps natural-language standards into machine-readable predicates subject to engineering verification.

Beyond semantic challenges, a trust gap persists due to the conflict between the probabilistic nature of AI and the deterministic requirements of safety circuits. On cost-effective edge platforms like the Rockchip RK3588, software-level latencies often introduce unpredictable ``tail latencies'' that can exceed strict safety budgets. To address this, we propose a symmetric \textbf{Dual-Modular Redundancy (DMR)} architecture with a \textbf{Parallel Independent Execution (PIE)} mechanism. By integrating Analog-to-Digital Converter (ADC)-based hardware probing and Universal Asynchronous Receiver-Transmitter (UART) heartbeats, the architecture ensures safety continuity and self-healing through rapid fault detection and automatic recovery \cite{Fault_tolerance}. Under the assumed single-fault condition, this design allows the peer node to maintain protective operation while the faulty node is detected and recovered, thereby reducing the impact of stochastic software uncertainties on the safety function. Finally, we demonstrate the framework through a low-latency perception-compute-control (PPC) architecture deployed in a real-world closed loop, providing a practical implementation path toward \textbf{ISO 13849 Category 3 and Performance Level d (PL d)} on cost-effective edge hardware.

The main contributions of this work are summarized as follows:
\begin{itemize}
    \item \textbf{A Multi-Tiered Safety Semantic Hierarchy for HRC:} We move beyond 1D distance-based metrics to propose a 3D safety definition encompassing Human-Robot safety protection, anomalous target identification, and anomalous behavior recognition. This hierarchy provides a more nuanced and context-aware interpretation of safety compared to traditional binary detection.
    
    \item \textbf{An LLM-Guided Automated Formalization Pipeline:} We develop a semi-automated pipeline that utilizes LLMs to assist in grounding safety definitions by parsing natural-language safety standards into mathematical predicates. This improves traceability from regulatory text to machine execution and reduces manual effort in safety engineering.
    
    \item \textbf{A Redundant Architecture with Safety Continuity and Self-Healing:} We engineer a symmetric DMR system utilizing a Parallel Independent Execution mechanism. By integrating ADC-based hardware probing and UART heartbeat, the architecture enables rapid fault detection and automatic recovery, ensuring the peer node remains fully operational during a restart.
    
    \item \textbf{A Low-Latency Perception-Compute-Control Architecture:} We demonstrate a low-latency architecture enabling direct end-to-end perception-to-control execution in a closed loop, and validate its feasibility through real-world deployment on RK3588 hardware. Using the \textit{Human-Robot Safety Protection} task as the primary benchmark, we show that the system provides a practical implementation path (as illustrated in Fig.~\ref{forward}) toward ISO 13849 Category 3 and Performance Level d (PL d) requirements.
\end{itemize}

\section{Related Work}
This section reviews research progress related to the work in this paper, focusing on three core dimensions: first, the evolution of safety definitions in HRC; second, the use of LLMs to automatically convert natural language (NL) safety specifications into formal expressions; and third, research on redundant fault-tolerant architectures and real-time execution for safety-critical AI perception systems. Existing research has explored various facets of these dimensions, such as parsing regulations or requirements into logical forms to provide formal support for safety verification \cite{Meng_LLM2026}. Other studies focus on improving the reliability and fault tolerance of perception systems through parallel multi-model and multi-sensor operation and analyzing their real-time performance \cite{Reliability_in_auto_vehicles}. Although these methods each have their own advantages and disadvantages, they often rely on heavy manual involvement or incur high computational overhead. In contrast, our work aims to connect three aspects that are usually studied separately: multidimensional safety definition in HRC, LLM-assisted formalization of regulatory text, and deterministic redundant execution on cost-effective edge hardware. Rather than optimizing only one of these components, we study how they can be composed into a single framework that links semantic safety requirements to runtime safety actions.

\subsection{Safety Definition in HRC}
Existing studies on safety in HRC largely define safety from a single perspective. Vision-based safety systems formulate safety in terms of spatial separation, relying on human detection and distance monitoring to enforce protective actions\cite{vision_based}. Similarly, dynamic safety zone approaches extend this paradigm by adapting safety boundaries according to robot motion and environmental context, yet their definition of safety remains fundamentally geometry-driven\cite{Dynamic_Safety_Zones}. These methods are effective for real-time enforcement but are primarily limited to spatial constraints.

On the other hand, anomaly detection methods focus on identifying deviations in robotic processes, such as failures in assembly or manipulation tasks\cite{Grambow_2026_anomaly_in_HRC}. While these approaches enhance system robustness, they typically treat anomalies as isolated events and do not integrate them into a unified safety framework for HRC. In contrast, our definition combines spatial safety, target integrity, and behavioral dynamics into a multi-dimensional formulation, enabling a more comprehensive and formally verifiable representation of safety beyond the scope of existing methods.

\subsection{LLM-Based Specification Formalization and Automation}
In recent years, several works have explored the use of LLMs to convert NL safety/policy texts into formal rules or logical expressions. One approach is the interactive translation of NL requirements into logical forms or executable constraints. For instance, Meng et al. \cite{Meng_LLM2026} proposed an LLM-assisted tool to interactively convert unstructured requirement text into logical forms, capturing the meaning of requirements with high fidelity while avoiding ambiguity. The RAFT framework \cite{xue2026explicatingtacitregulatoryknowledge} proposed by Xue et al. adopts a multi-round prompt strategy to ``explicitly extract'' regulatory knowledge from LLMs, generating verifiable formal requirements and test constraints, which significantly reduces expert modeling costs. 

Beyond requirement parsing, recent research emphasizes grounding LLM-generated plans in formal logic to ensure the safety of robot control tasks. Obi et al \cite{obi2026preexecutionsafetygate}. designed the SafeGate system, drawing on the ISO 13482 standard to extract safety constraints from NL instructions and checking them via SMT (Satisfiability Modulo Theories) to block dangerous commands. Similarly, Wu et al. \cite{wu2026grounding} proposed a hybrid architecture for trustworthy embodied AI, which grounds generative planners in verifiable logic to bridge the gap between high-level reasoning and formal safety execution. Furthermore, Yang et al. \cite{Yang_Safety_Chip} proposed the ``Safety Chip,'' which uses prompts to translate NL safety instructions into Linear Temporal Logic (LTL) formulas to automatically verify whether LLM-generated plans satisfy formal safety standards (compatible with ISO 61508).

Overall, the aforementioned methods confirm the feasibility of using LLMs to transform textual specifications into formal rules. For example, Endres et al. \cite{Endres} studied the conversion from NL to formal postconditions and found that LLMs exhibit high accuracy in transforming NL intent into program-verifiable assertions. These works indicate that LLMs potentially ``encode'' specification knowledge and, through appropriate prompting, can output machine-readable safety logic. However, they also possess certain limitations: many solutions still require manual involvement in review and correction, and the stability of generated results can be limited by prompt design \cite{xue2026explicatingtacitregulatoryknowledge, datla2025executablegovernanceaitranslating}. Our work builds on these ideas but further emphasizes a top-down formalization pipeline. By combining multi-round prompts with automated induction, we strive to reduce manual intervention and make the conversion from ISO text to Boolean safety conditions more controllable and verifiable.

\subsection{Redundant Architectures and Real-Time Guarantees for Safety Perception}
Safety-critical systems typically enhance reliability through redundant design. Many studies in fields such as autonomous driving and robotics have introduced multi-model or multi-sensor parallel structures. Pan et al. \cite{Reliability_in_auto_vehicles} pointed out that to ensure the reliability of perception systems, autonomous driving often employs parallel fusion of multiple sensors, such as cameras and LiDAR, to provide redundant perception. Other work focuses on hardware-level fault tolerance, such as the FPGA-based ``Auto-Healer'' self-healing architecture \cite{Auto_Healer} proposed by Suvizi et al. for perception chains in ADS. This architecture utilizes DMR to back up deep learning inference and implements hardware-level error detection and recovery with almost no increase in system latency. While these studies demonstrate that parallel redundancy is a vital means of improving perception safety, it typically brings additional computational and synchronization overhead.

\section{Safety Agent Framework}

Existing HRC safety systems often lack semantic richness, while AI perception systems often lack deterministic safety realization. To bridge this gap, we present a safety definition for HRC together with a top-down, LLM-guided framework that links industrial safety regulations, perception outputs, and runtime safety execution within an integrated PPC architecture on a redundant hardware platform. The framework connects abstract natural-language requirements to deterministic hardware behavior through a multi-layer mapping process.

The proposed framework establishes a hierarchical flow consisting of three primary stages: (i) LLM-based semantic formalization, (ii) multidimensional alignment of perception results, and (iii) redundant hardware execution with cross-layer monitoring. As illustrated in Fig.~\ref{framework}, the paradigm is instantiated using a collaborative robot scenario, where ISO 13849-1 standards are mapped onto a dual-RK3588 hardware platform for Human-Robot Safety Protection.

\subsection{Safety Definition in HRC Scenario}
In this work, we propose a comprehensive safety definition for HRC by integrating three distinct sensing tasks. Rather than relying solely on distance-based metrics \cite{SSM}, this definition addresses safety from the perspectives of spatial boundaries, target integrity, and behavioral dynamics to provide all-around protection.

The first task is \textbf{Human-Robot Safety Protection}, which focuses on the spatial dimension of safety. By projecting perception results into the 3D robotic coordinate system, the system monitors predefined safety zones to ensure immediate adherence to ISO 13849-1 standards upon any physical breach. The second task is \textbf{Anomalous Target Identification}, which monitors the integrity of entities within the workspace. By identifying anomalous targets that deviate from standard operational profiles, the system can initiate specific safety protocols for unidentified or unauthorized objects. The third task is \textbf{Anomalous Behavior Recognition} \cite{Morais_2019_CVPR_skeleton}, which evaluates the temporal aspect of safety. This task analyzes human motion sequences to identify hazardous patterns—such as sudden falls or erratic movements toward the robot—enabling proactive safeguarding before a physical collision occurs. 

By defining safety through these three tasks, the framework moves beyond simple binary detection. This multi-dimensional approach ensures that the LLM-based formalization pipeline can map NL regulations into a set of robust, verifiable safety predicates.

\begin{figure*}[htbp]
\centerline{\includegraphics[width=1.0\textwidth]{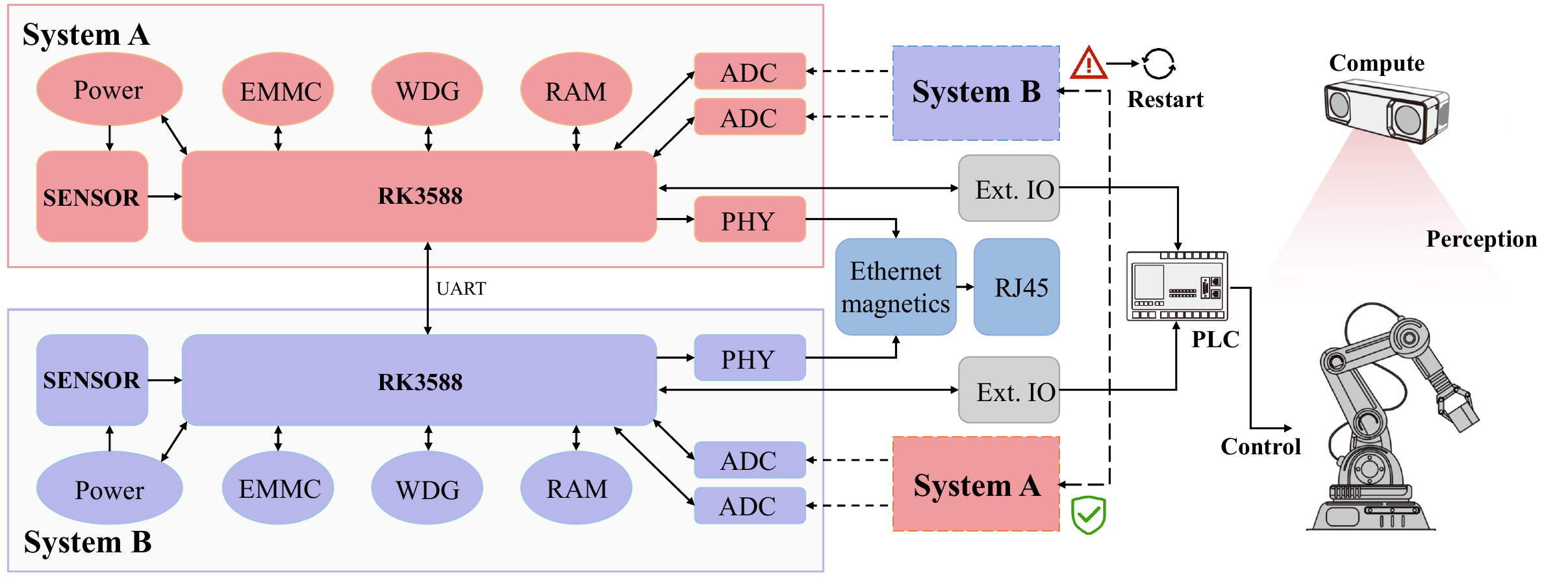}}
\caption{Hardware architecture of the integrated PPC system. The design features a symmetric dual-node redundancy centered on the Rockchip RK3588 SoC, facilitating an end-to-end pipeline from raw sensor acquisition to deterministic hardware execution via Ext. IO.}
\label{architecture}
\end{figure*}

\subsection{LLM-Based Safety Formalization}

\begin{algorithm}[t]
\caption{LLM-Guided Safety Formalization}
\label{alg:llm_formalization}
\begin{algorithmic}[1]
\REQUIRE Persona $\mathcal{P}$, Standard $\mathcal{S}$ (ISO 13849), Context $\mathcal{C}$
\ENSURE Predicate $S_{safe}$, DC Requirement $\mathcal{R}_{DC}$

\STATE $\mathcal{Q} \leftarrow \text{ConstructPrompt}(\mathcal{P}, \mathcal{S}, \mathcal{C})$ \COMMENT{Expert persona setup}
\STATE $\mathcal{L} \leftarrow \text{LLM\_Inference}(\mathcal{Q})$ \COMMENT{Knowledge-based reasoning}

\STATE $\{v_{max}, T_{stop}, d_{brake}, DC\} \leftarrow \text{ParseVars}(\mathcal{L})$ \COMMENT{Extract key variables}

\STATE $d_{offset} \leftarrow 1.5 \cdot d_{brake}$ \COMMENT{Apply conservative safety margin}
\STATE $S_{safe}(t) \equiv  d(t) \geq v_{max} \cdot T_{stop} + d_{offset}$

\STATE $\mathcal{R}_{DC} \leftarrow \text{CheckCompliance}(\mathcal{L}, \text{Cat. 3})$ \COMMENT{Verify fault tolerance}
\IF{$\mathcal{R}_{DC} \text{ is Valid}$}
    \STATE \textbf{return} $S_{safe}, \mathcal{R}_{DC}$
\ELSE
    \STATE \textbf{Abort} ``Safety Definition Inconsistent''
\ENDIF
\end{algorithmic}
\end{algorithm}

The bridge between regulatory requirements and technical implementation is often hindered by the inherent ambiguity of NL in industrial manuals \cite{NL_ambiguity}. Traditional manual interpretation of standards like ISO 13849-1 is not only time-consuming but also prone to subjective bias, which may lead to inconsistent safety configurations. To address this, we utilize LLMs to perform automated semantic parsing of the ``Safety of Machinery'' standard (see Algorithm~\ref{alg:llm_formalization}). 

The formalization strategy centers on a specialized prompt structure $\mathcal{Q}$ that defines the LLM’s persona $\mathcal{P}$ as a \textbf{Senior Functional Safety Expert} and the operational context $\mathcal{C}$ of the HRC. To ensure formal rigor, the prompt incorporates explicit verification criteria that guide the LLM's engineering interpretation of the standard $\mathcal{S}$. This structured approach enables a precise safety definition by grounding abstract regulations into the semantic parsing result $\mathcal{L}$. Under this persona, the LLM identifies critical deterministic variables—such as maximum velocity $v_{max}$, signal latency $T_{stop}$, braking distance $d_{brake}$, and Diagnostic Coverage (DC) levels $\mathcal{R}_{DC}$. These variables are then synthesized into formal predicates, providing a structured mathematical representation of the safety intent expressed in the manual.

For HRC, the primary safety objective extracted by the LLM is to ensure that the robot reaches a safe state before an operator crosses the remaining separation distance, even in the event of a single point failure. This is formalized as the safety predicate $S_{safe}$:
\begin{equation}
S_{safe}(t) \iff d(t) \geq v_{max} \cdot T_{stop} + d_{offset}
\end{equation}
where $d(t)$ is the instantaneous separation distance at time $t$, $v_{max}$ is the maximum relative velocity, $T_{stop}$ is the total signal latency, and $d_{offset}$ is a conservative constant representing the braking distance (e.g., set to 1.5 times the robot's maximum braking distance $d_{brake}$). Furthermore, the framework extracts requirements for diagnostic coverage (DC), stipulating that dangerous failures must be detected before or at the next demand of the safety function. In the current implementation, the LLM is used during the design stage to assist requirement formalization, while the runtime safety decision is executed deterministically by the edge platform.

\subsection{Multidimensional Perception Alignment}
The formalized predicate is not used directly on raw perception outputs. Instead, the detected visual observations are first aligned with the physical workspace and the system timing model. At runtime, the safety decision is obtained by evaluating the aligned distance estimate and bounded execution latency against $S_{safe}(t)$, and the result is then mapped to zone-based control actions.

While LLM-based formalization provides a mathematical safety objective, a significant gap remains between the probabilistic outputs of AI models (e.g., bounding boxes with confidence scores) and the deterministic requirements of industrial safety. To bridge this gap, the framework must map these high-level perceptions into a structured format that the safety formula can evaluate. This is achieved through multidimensional perception alignment, which synchronizes the AI's output with the physical world across spatial and temporal dimensions.

\subsubsection{Spatial alignment}
The necessity of spatial alignment stems from the fact that raw pixel-level detections lack physical scale and orientation relative to the robot's kinematics. To make the detection actionable, the perception bounding boxes are projected into the 3D robotic coordinate system to calculate the real-world distance $d(t)$. Based on the formalized safety predicate, the workspace is partitioned into three physical zones to provide a graded response to human-robot safety protection, where $d_{min}$ is the pre-defined safety distance:
\begin{itemize}
    \item \textbf{Stop Zone:} $d(t) < d_{min}$. Detection of any human pixels within this zone triggers an immediate Category 1 stop to prevent imminent collision.
    \item \textbf{Warning Zone:} $d(t) \geq d_{min}$ but within a safety margin. The robot operates at a reduced collaborative speed, balancing productivity with safety.
    \item \textbf{Safe Zone:} Sufficient separation exists, allowing the robot to operate at full rated speed without intervention.
\end{itemize}

\subsubsection{Temporal alignment}
Beyond spatial accuracy, functional safety is inherently time-critical. A delay in the perception-to-action pipeline can render a theoretically safe distance insufficient in practice. Therefore, the framework performs temporal alignment by decomposing the signal perception-to-safety-output latency budget, denoted here as $T_{stop}$, into its constituent latencies:
\begin{equation}
T_{stop} = t_{perc} + t_{infer} + t_{post}
\end{equation}
In this formulation, $t_{perc}$ represents the perception acquisition latency (including camera exposure and image pre-processing), $t_{infer}$ denotes the NPU-based inference latency which is the primary subject of our Worst-Case Execution Time (WCET) \cite{WCET} analysis, and $t_{post}$ accounts for the post-processing covering coordinate transformation and safety predicate evaluation. In this work, the mechanical braking characteristic is absorbed into the conservative distance margin $d_{offset}$, while the measured $T_{stop}$ focuses on the signal perception, inference, and decision path.
By bounding the WCET of the entire perception-to-control pipeline, the framework ensures that real-time execution strictly adheres to the temporal constraints extracted from the ISO 13849-1 standard, transforming probabilistic AI performance into guaranteed safety timing.

\subsection{The End-to-End Perception-Compute-Control Architecture}
The framework is architected as a direct PPC loop, representing an end-to-end safety system that bridges raw sensing and hardware execution in a unified pipeline. Unlike conventional safety systems that rely on complex intermediate steps or multi-stage software arbitration, our approach maps visual data directly to safety-critical control signals.
As shown in Fig.~\ref{architecture}, this end-to-end mapping reconciles deep-learning stochasticity with deterministic safety constraints. By bridging the industrial Programmable Logic Controller (PLC) bus with edge robotics protocols, the system achieves a seamless link between AI-based inference and low-level hardware execution.

\subsubsection{Heterogeneous computing node design}
To ensure sufficient computational power for complex AI models without sacrificing the determinism required for industrial safety, each subsystem is designed as an autonomous computing unit centered around a Rockchip RK3588 SoC. This heterogeneous platform integrates an octa-core CPU and a 6 TOPS NPU, providing the requisite computational throughput for concurrent execution of the PPC pipeline and real-time safety analysis. Within each node, the hardware is configured to support high-integrity execution: dedicated LPDDR4x RAM and eMMC storage provide strict operational isolation to prevent common-cause failures; an independent hardware Watchdog Timer (WDG) monitors local liveness to trigger safety states in case of software deadlocks; and a separate power domain ensures physical fault containment. This self-contained design establishes each SoC as a independent unit, forming the reliable foundation for the dual-channel safeguarding mechanism.

\subsubsection{Parallel independent execution}
Building upon the isolated hardware nodes, the overall system architecture coordinates these units through a PIE paradigm to satisfy stringent real-time constraints. While each node processes raw visual streams and generates safety control signals independently to avoid synchronization-induced jitter, the architecture incorporates a mutual supervision mechanism to guarantee system-wide integrity and end-to-end continuity. Specifically, the two nodes are interconnected via a UART interface for low-level status heartbeats. More importantly, as illustrated in the cross-connection in Fig.~\ref{architecture}, a hardware-level cross-monitoring layer is implemented using ADC. Each node utilizes these ADCs to monitor the electrical stability and safety signal integrity of its redundant peer. This decoupled yet supervised approach ensures that any hardware degradation or logic inconsistency in one channel is immediately detected by the other, effectively transforming high-performance AI perception into a deterministic, dual-channel safety output that interfaces directly with industrial PLC and robotics protocols.

\begin{algorithm}[t]
\caption{Reliable Safety Execution Loop}
\label{alg:software_algorithm}
\begin{algorithmic}[1]
\WHILE{System Active}
    \STATE $\mathcal{F}_{new} \leftarrow$ \textit{Capture \& Resize}();
    \STATE \textbf{LIFO\_Update}($\mathcal{Q}, \mathcal{F}_{new}$); \COMMENT{Ensuring data freshness}

    \STATE \textbf{Parallel\_Inference}:
    \STATE $\mathcal{B}_{boxes} \leftarrow$ \textbf{NPU\_Invoke}($\mathcal{C}_{1 \dots n}, \mathcal{Q}.pop()$);
    
    \STATE \textbf{Safety Guard}:
    \IF{\textbf{ADC\_Monitor}() == \textit{Stable} \AND $t_{exec} < T_{limit}$}
        \STATE \textbf{Output}(\textit{Safety\_Logic}($\mathcal{B}_{boxes}$));
        \STATE \textbf{Heartbeat}(UART);
    \ELSE
        \STATE \textbf{Output}(\textit{EMERGENCY\_STOP});
    \ENDIF
\ENDWHILE
\end{algorithmic}
\end{algorithm}

\subsubsection{Optimized safety loop}

To fully exploit the hardware performance of the RK3588 while minimizing safety-critical latency, we implement an asynchronous execution flow (see Algorithm~\ref{alg:software_algorithm}). The software stack orchestrates parallel RKNN contexts $\mathcal{C}_{1 \dots n}$ to saturate the NPU's localized cores, maximizing throughput. To eliminate the ``stale data'' bottleneck—a common risk in deep-learning buffers—we employ an asynchronous Last-In, First-Out (LIFO) queue $\mathcal{Q}$ with a depth of one. This forces the system to drop outdated frames and prioritize the most recent perception data $\mathcal{F}_{new}$, ensuring that safety decisions based on detection results $\mathcal{B}_{boxes}$ are made with minimal temporal staleness. The Safety Guard mechanism verifies that the measured execution time $t_{exec}$ remains within a deterministic temporal boundary $T_{limit}$. Coupled with NEAREST scaling and multi-threading, this orchestration guarantees that the framework operates within a deterministic time window, providing a hard temporal boundary for the safety agent.

\subsubsection{Safety logic and cross-layer monitoring}
The system generates safety predicates by evaluating spatial criteria including centroid trajectories, ground-point projections, and area-based intersections. This logic is protected by a cross-layer monitoring strategy: while the UART heartbeat tracks software vitality, the ADC hardware probing samples critical voltage rails to detect hardware-level brownouts or NPU hangs that software watchdogs might miss, ensuring a fail-safe transition in any failure mode, maintaining integrity of the end-to-end control signal.

\section{Experimental Evaluation}
In this section, we evaluate the proposed LLM-based safety framework across three hierarchical tasks: \textbf{Human-Robot Safety Protection} (Task 1), \textbf{Anomalous Target Identification} (Task 2), and \textbf{Anomalous Behavior Recognition} (Task 3). The evaluation focuses on three primary aspects: (1) the real-time determinism of Task 1 on edge hardware, (2) the diagnostic coverage of the hardware-software redundancy, and (3) the functional feasibility and semantic extensibility of the framework. 

\subsection{Experimental Setup}
\textbf{Hardware Configuration:} The framework is deployed on a heterogeneous edge platform consisting of two symmetric Rockchip RK3588 SoCs. Each SoC features an octa-core CPU (Cortex-A76/A55) and a 6 TOPS NPU. The systems are interconnected via high-speed UART for heartbeats and an integrated 12-bit ADC for cross-layer voltage and current monitoring. For advanced semantic tasks, a host-side workstation is utilized to establish performance baselines.

\textbf{Software Environment:} We utilize a single-shot safety agent, optimized via the RKNN-Toolkit2 for NPU acceleration. The software stack is implemented in C++17 for deterministic execution. The Human-Robot safety Protection is verified using a custom Human-Robot Interaction (HRI) dataset, which includes various industrial scenarios such as HRC, non-standard industrial tools (anomalous targets), and prohibited operator behaviors.

\textbf{Safety Evaluation Methodology:} In practical industrial deployments, the required safety separation distance is a function of both mechanical inertia and system response time. Since the mechanical braking distance is typically a fixed characteristic, the Safety Integrity of Task 1 depends primarily on its Response Latency Determinism. Therefore, we focus our edge-side evaluation on the End-to-End Signal Latency ($T_{stop}$). Furthermore, to evaluate the framework's breadth, \textbf{Anomalous Target Identification (Task 2)} and \textbf{Anomalous Behavior Recognition (Task 3)} are assessed using the \textbf{Area Under Curve (AUC)} metric. This dual-track methodology ensures that the robot can reach a safe state before contact (Task 1) while possessing the intelligence to identify complex hazards (Tasks 2 and 3) that exceed simple spatial constraints.

\begin{figure*}[tbp]
\centerline{\includegraphics[width=\textwidth]{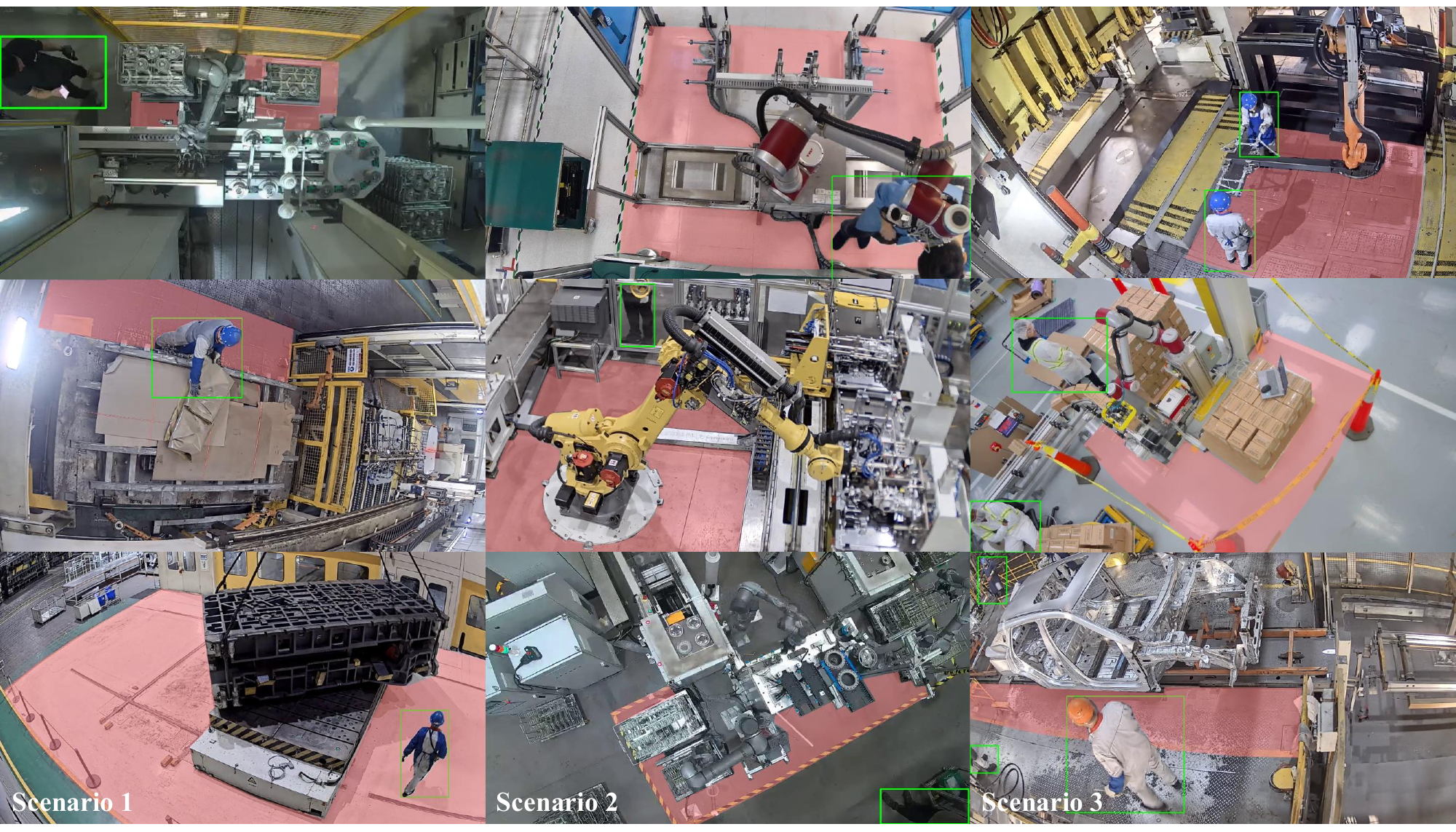}}
\caption{Visual representation of the three experimental scenarios utilized in the performance evaluation, with corresponding data metrics detailed in Table~\ref{tab:multi_scenario}. \textbf{Red floor regions} denote pre-defined danger zones ($R_{ROI}$), and \textbf{green bounding boxes} represent detected human subjects. (Left) Scenario 1: A standard industrial workspace with a single worker under clear overhead visibility. (Middle) Scenario 2: A challenging environment where the worker is partially obscured. (Right) Scenario 3: A complex spatial layout featuring two or more workers in a multi-target configuration.}
\label{scenario}
\end{figure*}

\subsection{Temporal Performance and WCET Analysis}

\begin{table}[t]
\centering
\caption{Latency Breakdown and WCET Analysis across Multiple Scenarios (ms)}
\label{tab:multi_scenario}
\begin{tabular}{l l c c c}
\toprule
\textbf{Scenario} & \textbf{Component} & \textbf{Average} & \textbf{WCET} & \textbf{Std. Dev.} \\ 
\midrule

\multirow{4}{*}{Scenario 1} 
& $t_{perc}$  & 7.50 & 24.60 & 2.18 \\
& $t_{infer}$ & 25.05 & 39.11 & 0.51 \\
& $t_{post}$  & 2.41 & 11.95 & 1.16 \\
\cmidrule{2-5}
& \textbf{$T_{stop}$} & \textbf{34.96} & \textbf{57.66} & \textbf{2.52} \\ 
\midrule

\multirow{4}{*}{Scenario 2} 
& $t_{perc}$  & 7.15 & 21.13 & 2.02 \\
& $t_{infer}$ & 24.96 & 37.52 & 0.64 \\
& $t_{post}$  & 2.87 & 10.73 & 1.08 \\
\cmidrule{2-5}
& \textbf{$T_{stop}$} & \textbf{34.98} & \textbf{52.60} & \textbf{2.38} \\ 
\midrule

\multirow{4}{*}{Scenario 3} 
& $t_{perc}$  & 8.43 & 25.03 & 2.97 \\
& $t_{infer}$ & 24.44 & 31.49 & 0.45 \\
& $t_{post}$  & 6.64 & 15.35 & 2.40 \\
\cmidrule{2-5}
& \textbf{$T_{stop}$} & \textbf{39.51} & \textbf{53.87} & \textbf{3.84} \\ 
\bottomrule
\end{tabular}
\end{table}

The real-time determinism of the perception pipeline is the cornerstone of functional safety. Unlike consumer-grade AI applications that prioritize average throughput, our framework emphasizes the \textbf{WCET} to ensure that safety predicates are evaluated within a strictly bounded temporal envelope.

\begin{table}[t]
\centering
\caption{Fault Injection and Recovery Analysis}
\label{tab:fault_injection_results}
\footnotesize
\begin{tabularx}{0.47\textwidth}{@{\extracolsep{\fill}}l l c c@{}}
\toprule
\textbf{Fault} & \textbf{Mechanism} & \textbf{$T_{det}$ (ms)} & \textbf{$T_{rec}$ (ms)} \\ \midrule
\textbf{Heartbeat Loss}     & SW Logic  & 51.87  & 39627.63 \\ \midrule
NPU Hang                    & SW Logic & 2.04   & 313.61   \\ \midrule
Power Brownout              & \textbf{ADC Probing} & 38.45  & 39546.52 \\ \midrule
Sensor Fault                & SW Logic & 2010.45 & 1236.17  \\ 
\bottomrule
\end{tabularx}
\end{table}

We performed extensive profiling of a lightweight safety agent over 10,000 inference cycles across three representative HRI scenarios illustrated in Fig.~\ref{scenario}: baseline visibility, partial occlusion, and multi-target interaction. As illustrated by the data in Table~\ref{tab:multi_scenario}, the lightweight nature of single-shot safety agent, combined with our INT8 quantization and memory-locking strategy, significantly suppresses execution jitter. Specifically, the NPU-based inference stage ($t_{infer}$) maintains an average latency of approximately 25 ms with a remarkably low standard deviation ($\sigma \le 0.64$ ms). The gap between the average $T_{stop}$ and its WCET is strictly maintained within 65\%, providing a predictable upper bound for the safety separation distance calculation. 

While the post-processing stage ($t_{post}$) exhibits a marginal increase in Scenario 3 (6.64 ms) due to the computational overhead of multi-target coordinate transformation and spatial verification, the jitter remains bounded through our deterministic C++ implementation and SCHED\_FIFO real-time task priority. By establishing these rigorous temporal bounds across diverse industrial conditions, we show that the system can reliably maintain the safety distance $S_{safe}(t)$ even under peak computational loads, effectively eliminating the risk of late-triggered braking.

\subsection{Fault Tolerance and Diagnostic Coverage}

The experimental results in Table~\ref{tab:fault_injection_results} validate the system's rapid diagnostic and recovery capabilities. Critical failures like the \textbf{NPU Hang} and \textbf{Power Brownout} are detected within 2.04 ms and 38.45 ms respectively, ensuring a high-speed response to hardware-level anomalies. While \textit{Heartbeat Loss} and \textit{Sensor Fault} exhibit longer $T_{det}$ values (51.87 ms and 2010.45 ms), these are intentional algorithmic thresholds—set at 50 ms and 2000 ms respectively—to filter transient noise and prevent false triggers.

Furthermore, the system implements an automatic recovery mechanism. Although hardware reboots for severe faults take approximately 40 seconds, safety is never compromised. Under the PIE architecture, the redundant node remains fully operational while its peer restarts. This non-stop perception ensures the system consistently satisfies ISO 13849 Category 3 requirements, maintaining safety continuity even during individual node recovery cycles.

\subsection{Feasibility Study: Advanced Semantic Integration}

\begin{table}[t]
\centering
\small
\caption{Feasibility Benchmarking of Advanced Semantic Tasks across Multiple Datasets}
\label{tab:auc_results_detailed}
\begin{tabular*}{0.47\textwidth}{@{\extracolsep{\fill}}llc@{}}
\toprule
\textbf{Task Category} & \textbf{Evaluation Dataset} & \textbf{AUC} \\ \midrule
\multirow{3}{*}{\shortstack[l]{\textbf{Task 2:}\\Anomalous Target\\Identification}} 
    & IPAD-R1\cite{IPAD}  & 0.84 \\
    & IPAD-R2\cite{IPAD}  & 0.75 \\
    & IPAD-R3\cite{IPAD}  & 0.45 \\
\cmidrule(lr){2-3}
    & \textit{Average}      & \textbf{0.68} \\ \midrule
\multirow{3}{*}{\shortstack[l]{\textbf{Task 3:}\\Anomalous Behavior\\Recognition}} 
    & HR-STC\cite{Shanghai_tech}   & 0.74 \\
    & HR-Ubnormal\cite{Acsintoae_CVPR_2022_UBnormal} & 0.64 \\
    & HR-Avenue\cite{avenue}  & 0.87 \\
\cmidrule(lr){2-3}
    & \textit{Average}      & \textbf{0.75} \\ \bottomrule
\end{tabular*}
\end{table}

To explore the framework's potential in handling complex safety logic beyond basic Human-Robot Safety Protection, we incorporate two representative safety scenarios as case studies: \textbf{Anomalous Target Identification} (Task 2) and \textbf{Anomalous Behavior Recognition} (Task 3). These tasks with their evaluation results detailed in Table~\ref{tab:auc_results_detailed}, are designed to assess the feasibility of integrating high-level semantic reasoning into the proposed LLM-guided pipeline. While these evaluations are currently conducted in a host-side environment to verify their discriminative power, the continuous scaling of on-device computing power and the advancement of hardware-specific NPU optimizations are expected to bridge this gap. Looking forward, these sophisticated modules are poised to be seamlessly integrated into the real-time edge execution loop, providing enhanced safety intelligence and robust protection for future industrial HRC applications.

\subsection{Discussion: Architectural Scalability and Future Integration}

The synthesis of edge-side latency data and host-side feasibility tests provides a preliminary roadmap for the system's evolution. The deterministic performance of Task 1 across varying scenarios confirms the reliability of the redundant hardware for immediate fail-safe deployment. 

Concurrently, the integration of Tasks 2 and 3 serves as a proof-of-concept for the framework's long-term scalability. The results indicate that the LLM-guided pipeline can effectively encapsulate complex safety semantics, even though the computational intensity of such high-dimensional features currently resides at the host level. As edge-AI accelerators (NPUs) continue to advance and models undergo further optimization through quantization, the architectural readiness of our PIE system allows for the seamless migration of these semantic modules, ensuring the system remains adaptable to the increasing complexity of HRC environments.

\section{Conclusion}

We introduce a dependable \textbf{safety agent} centered on \textbf{a multi-tiered safety definition} and an \textbf{LLM-guided formalization pipeline}, designed to bridge the gap between probabilistic AI perception and deterministic industrial safety requirements. By systematically parsing NL regulations into formal mathematical predicates, our pipeline provides a structured methodology for defining complex safety logics across multiple hierarchical tasks—ranging from basic Human-Robot Safety Protection to advanced anomalous behavior recognition. 

We validated the framework through a low-latency \textbf{Perception-Compute-Control} architecture inspired by ISO 13849-1 requirements on a redundant heterogeneous platform. The experimental results show that the proposed DMR architecture, coupled with the PIE mechanism, achieves bounded latency and strong diagnostic behavior under the evaluated workloads. Rather than claiming full certification, this work provides a practical implementation path for grounding LLM-assisted safety formalization in cost-effective hardware toward industrial safety deployment. This study establishes a practical benchmark for deploying versatile, ISO-compliant safety agents in next-generation industrial applications.


\bibliographystyle{ACM-Reference-Format}
\bibliography{sample-base}

@String{Computing = "Computing" }

@String{Computer = "{IEEE} Computer" }

@String{Springer = "Springer-Verlag" }

@article{Meng_LLM2026,
  title={Transforming Natural Language Requirements to Formalism Using LLMs},
  author={Meng, Baoluo and Lorch, Robert and Siu, Kit and Durling, Michael and Varanasi, Sarat Chandra and Paul, Saswata and Moitra, Abha},
  journal={Systems Engineering},
  volume={29},
  number={2},
  pages={195--204},
  year={2026},
  publisher={Wiley Online Library}
}

@article{xue2026explicatingtacitregulatoryknowledge,
  title={Explicating Tacit Regulatory Knowledge from LLMs to Auto-Formalize Requirements for Compliance Test Case Generation},
  author={Xue, Zhiyi and Chen, Xiaohong and Zhang, Min},
  journal={arXiv preprint arXiv:2601.09762},
  year={2026}
}

@article{obi2026preexecutionsafetygate,
  title={Pre-Execution Safety Gate \& Task Safety Contracts for LLM-Controlled Robot Systems},
  author={Obi, Ike and Venkatesh, Vishnunandan LN and Wang, Weizheng and Wang, Ruiqi and Suh, Dayoon and Amosa, Temitope I and Jo, Wonse and Min, Byung-Cheol},
  journal={arXiv preprint arXiv:2604.05427},
  year={2026}
}

@inproceedings{Yang_Safety_Chip,
  title={Plug in the safety chip: Enforcing constraints for llm-driven robot agents},
  author={Yang, Ziyi and Raman, Shreyas S and Shah, Ankit and Tellex, Stefanie},
  booktitle={2024 IEEE International Conference on Robotics and Automation (ICRA)},
  pages={14435--14442},
  year={2024},
  organization={IEEE}
}

@article{Endres,
  title={Can large language models transform natural language intent into formal method postconditions?},
  author={Endres, Madeline and Fakhoury, Sarah and Chakraborty, Saikat and Lahiri, Shuvendu K},
  journal={Proceedings of the ACM on Software Engineering},
  volume={1},
  number={FSE},
  pages={1889--1912},
  year={2024},
  publisher = {Association for Computing Machinery}
}

@inproceedings{Auto_Healer,
  title={Auto-Healer: Self-Healing Hardware for Perception Stage Faults in Autonomous Driving Systems},
  author={Suvizi, Ali and Venkataramani, Guru},
  booktitle={Proceedings of the 39th ACM International Conference on Supercomputing},
  pages={1064--1078},
  year={2025}
}

@article{Reliability_in_auto_vehicles,
  title={Reliability modeling for perception systems in autonomous vehicles: A recursive event-triggering point process approach},
  author={Pan, Fenglian and Zhang, Yinwei and Liu, Jian and Head, Larry and Elli, Maria and Alvarez, Ignacio},
  journal={Transportation Research Part C: Emerging Technologies},
  volume={169},
  pages={104868},
  year={2024},
  publisher={Elsevier}
}

@article{datla2025executablegovernanceaitranslating,
  title={Executable Governance for AI: Translating Policies into Rules Using LLMs},
  author={Datla, Gautam Varma and Vurity, Anudeep and Dash, Tejaswani and Ahmad, Tazeem and Adnan, Mohd and Rafi, Saima},
  journal={arXiv preprint arXiv:2512.04408},
  year={2025}
}

@inproceedings{wu2026grounding,
    title={Grounding Generative Planners in Verifiable Logic: A Hybrid Architecture for Trustworthy Embodied {AI}},
    author={Feiyu Wu and Xu Zheng and Yue Qu and Zhuocheng Wang and Zicheng Feng and HUI LI},
    booktitle={The Fourteenth International Conference on Learning Representations},
    year={2026},
    url={https://openreview.net/forum?id=wb05ver1k8}
}

@inproceedings{Acsintoae_CVPR_2022_UBnormal,
  title={Ubnormal: New benchmark for supervised open-set video anomaly detection},
  author={Acsintoae, Andra and Florescu, Andrei and Georgescu, Mariana-Iuliana and Mare, Tudor and Sumedrea, Paul and Ionescu, Radu Tudor and Khan, Fahad Shahbaz and Shah, Mubarak},
  booktitle={Proceedings of the IEEE/CVF conference on computer vision and pattern recognition},
  pages={20143--20153},
  year={2022}
}

@article{IPAD,
  title={Ipad: Industrial process anomaly detection dataset},
  author={Liu, Jinfan and Yan, Yichao and Li, Junjie and Zhao, Weiming and Chu, Pengzhi and Sheng, Xingdong and Liu, Yunhui and Yang, Xiaokang},
  journal={IEEE Transactions on Circuits and Systems for Video Technology},
  volume={35},
  number={1},
  pages={380--393},
  year={2024},
  publisher={IEEE}
}

@inproceedings{Shanghai_tech,
  title={A revisit of sparse coding based anomaly detection in stacked rnn framework},
  author={Luo, Weixin and Liu, Wen and Gao, Shenghua},
  booktitle={Proceedings of the IEEE international conference on computer vision},
  pages={341--349},
  year={2017}
}

@inproceedings{avenue,
  title={Abnormal event detection in videos using spatiotemporal autoencoder},
  author={Chong, Yong Shean and Tay, Yong Haur},
  booktitle={International symposium on neural networks},
  pages={189--196},
  year={2017},
  organization={Springer}
}

@article{Grambow_2026_anomaly_in_HRC,
  title={Anomaly detection for generic failure monitoring in robotic assembly, screwing and manipulation},
  author={Grambow, Niklas and Fenner, Lisa-Marie and Kempkes, Felipe and Hotz, Philip and Wan, Dingyuan and Kr{\"u}ger, J{\"o}rg and Haninger, Kevin},
  journal={IEEE Robotics and Automation Letters},
  year={2026},
  publisher={IEEE}
}

@inproceedings{vision_based,
  title={Vision-based safety system for barrierless human-robot collaboration},
  author={Amaya-Mej{\'\i}a, Lina Mar{\'\i}a and Duque-Su{\'a}rez, Nicol{\'a}s and Jaramillo-Ram{\'\i}rez, Daniel and Martinez, Carol},
  booktitle={2022 IEEE/RSJ International Conference on Intelligent Robots and Systems (IROS)},
  pages={7331--7336},
  year={2022},
  organization={IEEE}
}

@incollection{Dynamic_Safety_Zones,
  title={Dynamic Safety Zones in Human Robot Collaboration},
  author={Makris, Sotiris},
  booktitle={Cooperating Robots for Flexible Manufacturing},
  pages={271--287},
  year={2020},
  publisher={Springer}
}

@article{HRC_review,
  title={Reviewing human-robot collaboration in manufacturing: Opportunities and challenges in the context of industry 5.0},
  author={Dhanda, Mandeep and Rogers, Benedict Alexander and Hall, Stephanie and Dekoninck, Elies and Dhokia, Vimal},
  journal={Robotics and Computer-Integrated Manufacturing},
  volume={93},
  pages={102937},
  year={2025},
  publisher={Elsevier}
}

@article{industry5.0,
  title={Industry 5.0: Prospect and retrospect},
  author={Leng, Jiewu and Sha, Weinan and Wang, Baicun and Zheng, Pai and Zhuang, Cunbo and Liu, Qiang and Wuest, Thorsten and Mourtzis, Dimitris and Wang, Lihui},
  journal={Journal of Manufacturing Systems},
  volume={65},
  pages={279--295},
  year={2022},
  publisher={Elsevier}
}

@article{Fault_tolerance,
  title={Basic concepts and taxonomy of dependable and secure computing},
  author={Avizienis, Algirdas and Laprie, J-C and Randell, Brian and Landwehr, Carl},
  journal={IEEE transactions on dependable and secure computing},
  volume={1},
  number={1},
  pages={11--33},
  year={2004},
  publisher={IEEE}
}

@article{SSM,
  title={LiDAR-based maintenance of a safe distance between a human and a robot arm},
  author={Podgorelec, David and Uran, Suzana and Nerat, Andrej and Bratina, Bo{\v{z}}idar and Pe{\v{c}}nik, Sa{\v{s}}o and Dimec, Marjan and {\v{Z}}aberl, Franc and {\v{Z}}alik, Borut and {\v{S}}afari{\v{c}}, Riko},
  journal={Sensors},
  volume={23},
  number={9},
  pages={4305},
  year={2023},
  publisher={MDPI}
}

@inproceedings{Morais_2019_CVPR_skeleton,
  title={Learning regularity in skeleton trajectories for anomaly detection in videos},
  author={Morais, Romero and Le, Vuong and Tran, Truyen and Saha, Budhaditya and Mansour, Moussa and Venkatesh, Svetha},
  booktitle={Proceedings of the IEEE/CVF conference on computer vision and pattern recognition},
  pages={11996--12004},
  year={2019}
}

@incollection{NL_ambiguity,
  title={Ambiguity in requirements engineering: Towards a unifying framework},
  author={Gervasi, Vincenzo and Ferrari, Alessio and Zowghi, Didar and Spoletini, Paola},
  booktitle={From Software Engineering to Formal Methods and Tools, and Back: Essays Dedicated to Stefania Gnesi on the Occasion of Her 65th Birthday},
  pages={191--210},
  year={2019},
  publisher={Springer}
}

@article{WCET,
  title={The worst-case execution-time problem—overview of methods and survey of tools},
  author={Wilhelm, Reinhard and Engblom, Jakob and Ermedahl, Andreas and Holsti, Niklas and Thesing, Stephan and Whalley, David and Bernat, Guillem and Ferdinand, Christian and Heckmann, Reinhold and Mitra, Tulika and others},
  journal={ACM transactions on embedded computing systems (TECS)},
  volume={7},
  number={3},
  pages={1--53},
  year={2008},
  publisher={Association for Computing Machinery}
}

@techreport{iso13849,
  author      = {{International Organization for Standardization}},
  title       = {{ISO 13849-1:2023} Safety of Machinery --- Safety-related Parts of Control Systems --- Part 1: General Principles for Design},
  institution = {International Organization for Standardization},
  type        = {Standard},
  number      = {ISO 13849-1:2023},
  year        = {2023}
}

@article{rosenstrauch2018human,
  title={Human robot collaboration-using kinect v2 for iso/ts 15066 speed and separation monitoring},
  author={Rosenstrauch, Martin J and Pannen, Tessa J and Kr{\"u}ger, J{\"o}rg},
  journal={Procedia Cirp},
  volume={76},
  pages={183--186},
  year={2018},
  publisher={Elsevier}
}

@article{brunke2025semantically,
  title={Semantically safe robot manipulation: From semantic scene understanding to motion safeguards},
  author={Brunke, Lukas and Zhang, Yanni and R{\"o}mer, Ralf and Naimer, Jack and Staykov, Nikola and Zhou, Siqi and Schoellig, Angela P},
  journal={IEEE Robotics and Automation Letters},
  year={2025},
  publisher={IEEE}
}

@ArtifactSoftware{R,
    title = {R: A Language and Environment for Statistical Computing},
    author = {{R Core Team}},
    organization = {R Foundation for Statistical Computing},
    address = {Vienna, Austria},
    year = {2019},
    url = {https://www.R-project.org/},
}

\end{document}